\documentclass[12pt]{article}
\usepackage{lingmacros}
\usepackage[english]{babel}
\usepackage{graphicx}

\large \baselineskip = 20 true pt

\begin{document}
\begin{center}
{\Large \bf The algorithm of formation of a training set for an artificial neural network for image segmentation } \vspace{0.5cm}
\end{center}

\begin{center}
S.V. Belim, S.B. Larionov \\
Dostoevsky Omsk State University, Omsk, Russia
 \vspace{0.5cm}
\end{center}

\begin{center}
{\bf Abstract}
\end{center}

This article suggests an algorithm of formation a training set for artificial neural network in case of image segmentation. The distinctive feature of this algorithm is that it using only one image for segmentation. The segmentation performs using three-layer perceptron. The main method of the segmentation is a method of region growing. Neural network is using for get a decision to include pixel into an area or not. Impulse noise is using for generation of a training set. Pixels damaged by noise are not related to the same region. Suggested method has been tested with help of computer experiment in automatic and interactive modes.


\section{Introduction}
Image segmentation is one of the variants of the task of clustering a set of pixels on the basis of color affinity. It should be pointed out that the segmentation problem does not have an exact solution and the effectiveness of the decision is usually evaluated on the basis of comparison with the results obtained by experts in manual mode. That means that the effectiveness of the segmentation cannot be unambiguous. One of the approaches actively developing in recent few years is the use of artificial neural networks. However, this approach faces the problem of the formation of a learning set. If there is a series of similar images, for example, with medical information, then the problem is solved quite simply. But if the task is to segment one single image, which has no analogues, then the traditional approaches face unsolvable difficulties.
To solve the image segmentation problem, various artificial neural networks were used. There were built algorithms based on the multilayer perceptron \cite{b1}, as well as its modifications using cross entropy  \cite{b2}, genetic algorithms  \cite{b3, b4}, regions growing  \cite{b5, b6}, minimal difference ratio  \cite{b7}, wavelet decomposition of the original image  \cite{b8}. Additionally,  multilayer perceptron was used in conjunction with other segmentation algorithms, for example, the k-means method  \cite{b9}. A number of algorithms were also developed on the basis of Kohonen self-organizing maps  \cite{b10, b11, b12}. As well as for a multilayer perceptron, there are some algorithms were proposed to combine Kohonen maps with other  segmentation algorithms – the hybrid genetic algorithm  \cite{b13}, the k-means method  \cite{b14}, the growing of the regions  \cite{b15}.
All the algorithms listed above has specialized nature and oriented to images of a certain type. This specialization is necessary for the formation of a learning set. Neural networks trained on the one type of images do not work for other types of images.
In this paper, we propose an algorithm for forming a learning set in a segmentation problem that is oriented to one particular image and does not require an additional set of similar images.

\section{The formulation of the problem and the solving method}

The image segmentation algorithm will be based on the method of growing areas, which has proved to be very useful when using graph representation of the image \cite{b16, b17}. Growing a segment starts from a single point selected randomly on the raw portion of the image. At the initial point in time, the whole image is considered unprocessed. All points attached to one of the segments are considered to be the processed part of the image. Segment growing occurs by attaching new points to it. The algorithm traverses the boundary points of the already formed part of the segment and pairwise processes them with neighboring points that are not part of the segment. For each such pair, a decision is made whether to join or not to the segment. They join the segment of the point according to the color characteristics similar to those already included in it.
For each segment, a unique label is used, represented by a natural number. The raw part of the image is marked with a zero value. Thus, at each stage of the algorithm work, all the image points are associated with some labels. At the initial time all points have a zero mark. At the end of the algorithm all points are aligned with nonzero labels.
To make a decision on the similarity of two points, a three-layer perceptron was used. The points were compared by their color coordinates in the RGB model. Since the comparison occurs only for two points, the input layer of the perceptron has six neurons corresponding to the color coordinates of the points. The output layer contains two neurons, corresponding to a positive and negative decision about the similarity of points. In the framework of a computer experiment, it was determined that the algorithm's efficiency increases with the increase in the number of neurons in the hidden layer down to a value of 50, after which it remains unchanged. Therefore, in the future, a three-layer perceptron with fifty neurons in a hidden layer was used.
Assume that there is an image that has $N\times M$ pixels which in the RGB model can be represented with a three matrixes $||R_{ij}||$, $||G_{ij}||$, $||B_{ij}||$ ($i=1,...,N$; $j=1,...,M$), that corresponds to a color components red, green and blue respectively. Let’s introduce a notation $v_{ij}$ for a pixel with coordinates $(i,j)$:

\begin{equation}
v_{ij}.r=R_{ij},\ \ v_{ij}.g=G_{ij},\ \ v_{ij}.b=B_{ij}.
\end{equation}

For the matrix of marks let's introduce a notation $C_{ij}$ ($i=1,...N$; $j=1,...,M$). Let's use a notation $v_{ij}.c=C_{ij}$ for a mark of some specific pixel.

Let's inspect the algorithm of region growing. Assume that there are $k$ segments are found at the moment. Let's get a random point from the set of pixels with zero mark and demote it by $v_{ij}$. Changing the mark of this pixel: $v_{ij}.c=k+1$. Now let's run the following recursive algorithm:
\begin{enumerate}

\item{Assume the growing segment contain $r$ pixels $\{v_1, ..., v_r\}$.}

\item{Traverse all neighbours of the pixels with zero mark that are included into the segment.}

\item{Generating an input vector of the neural network using values of color components of a point of the segment $v$ and its nearest neighbour $v'$ that not included to the segment: \\
\begin{equation}
(v.r, v.g, v.b, v'.r, v'.g, v'.b).
\end{equation}
}

\item{Calculating output value of the neural network. If the decision is positive, change the mark of  point $v'$: $v'.c=k+1$. If the decision is negative, continue with next neighbour point that is not included to the segment.
}

\item{The algorithm stops when there are no positive decisions for all nearest neighbours.}

\end{enumerate}

Segmentation of the image is completed when there are no points with zero mark.
The algorithm will have a linear complexity such as it iterates all pixels that has not more then eight nearest neighbours.

The efficiency of the algorithm is significantly depends on the training of the neural network, which is determined
by a training set. Let's form a training set for a single image by using of artificially generated impulse noise. A pseudo-random generator sequences with uniform distribution will be used. Based on this generator, let's generate a new color coordinates for $p\%$ pixels of the original image. The coordinates
will be select randomly, and a new color will be choose so that it is significantly
differed from the original. Assume that the entire color palette contains $m$ colors. If in the original
image for a random pixel with coordinates $(i, j)$ one of the color coordinates
having the value $m_ {ij}> m / 2$, then let's select a new color from the interval $[0, m / 2)$. For
the case $m_ {ij} \leq m / 2$ the new color is chosen from the interval $(m / 2, m-1]$. As a result of this modification it can be considered that the modified pixel can not be combined into one segment with the nearest neighbours. Changed pixels playing the main role in the training set. As the computer experiment has shown, it is necessary to select the value $p \leq 10 \%$. To form a training set of a larger volume, it is necessary to repeat the procedure of impulse noise formation on the main image repeatedly. Neural network training has been performed using the method of back propagation of errors. The main disadvantage of this approach to the formation of the training set is that result set contain only negative decisions.

\section{Computer experiment}

During the computer experiment there is an image segmentation of the images with the depth of the palette of each color $m = 256$ has been performed. There are $10 \%$ of damaged pixels were generated and this generation has been performed 100 times. The algorithm was tested both in automatic and in interactive modes. In automatic mode, all segments of the image were determined. In the interactive mode, a single segment was searched that contained a user-defined point.

An example of how an algorithm works in automatic mode is shown in Figure 1, which shows
contours of the segments. Figure 2 shows the result of the algorithm in interactive mode.

\begin{figure}[ht]
\centering
\includegraphics[width=0.8\textwidth]{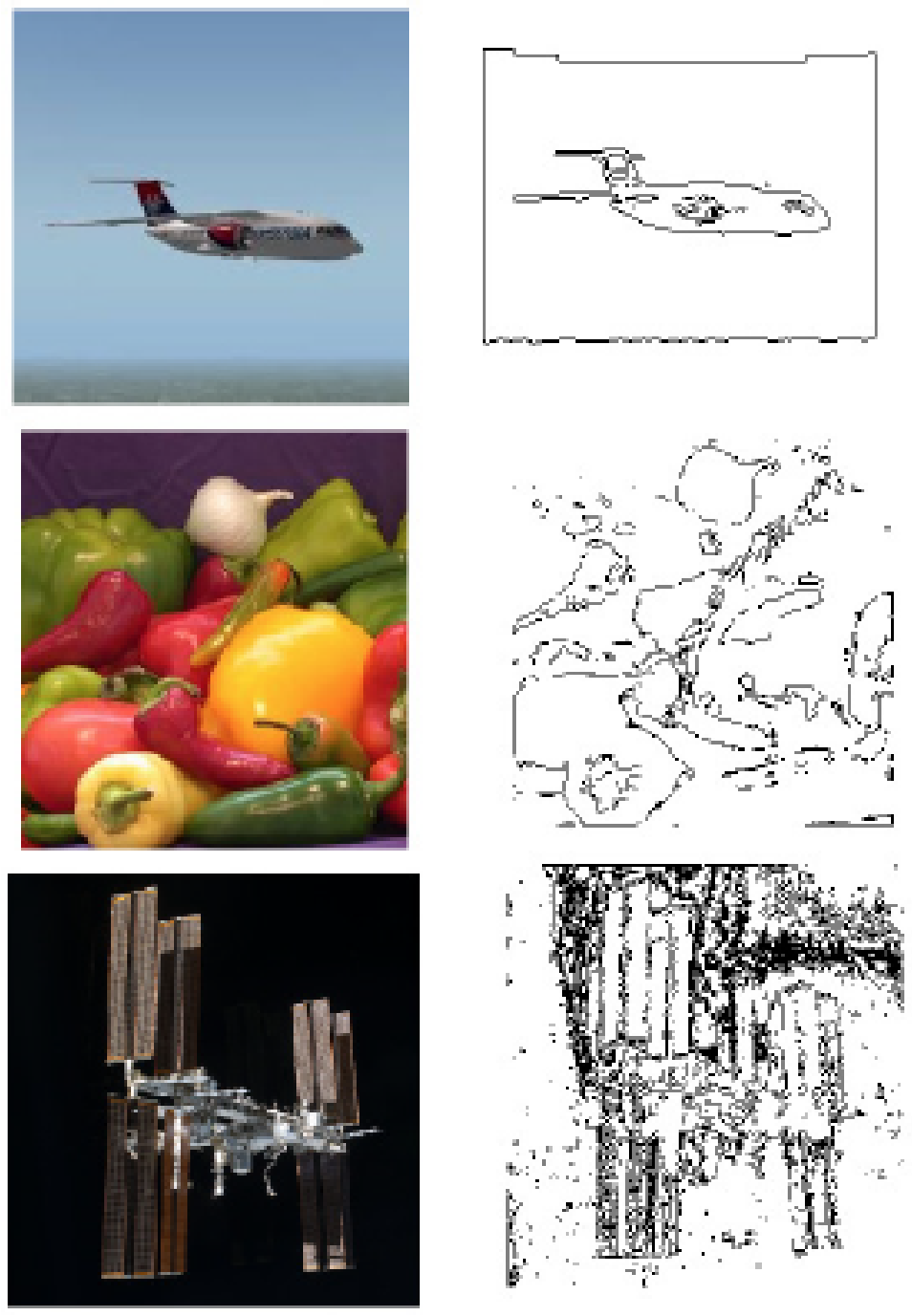}
\caption{Examples of automatic segmentation of the images.}
\label{fig1}
\end{figure}

\begin{figure}[ht]
\centering
\includegraphics[width=0.8\textwidth]{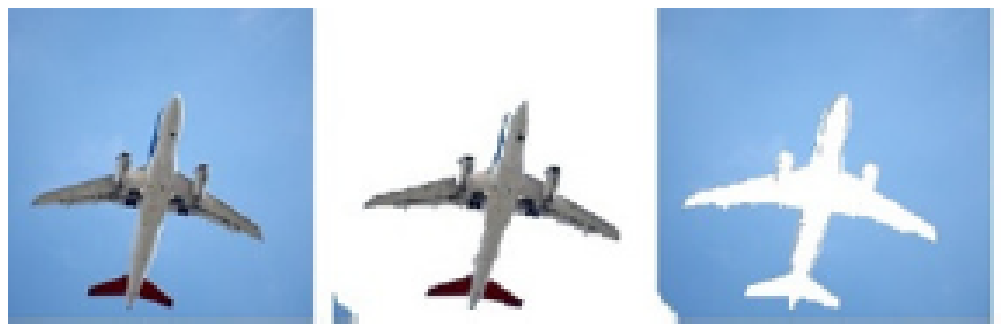}
\caption{Example of algorithm result using interactive mode on the image "Plane"}
\label{fig1}
\end{figure}

\section*{Conclusion}

Despite the existing shortcomings in the formation of the training set, the proposed algorithm allows the segmentation of a given image with a neural network without involving other images that are close in structure. A computer experiment showed that the proposed algorithm is most effective in the case of clear contours and segments of large size. Next, note that this property is inherent in all segmentation algorithms. However, it is necessary to note the stability of the proposed segmentation algorithm to the impulse noise, which is easily determined by the trained neural network, like segments with a size of one pixel. In the future, this approach can be used with other more efficient neural networks.

Also, the proposed algorithm has good speed due to the linear complexity. The main time is spent on the formation of training set and training of an artificial neural network. The segmentation process on an already trained neural network is much faster. The total processing time of the image size $256 \times 256$ pixels is about 40 seconds.


\begin{thebibliography}{00}

\bibitem{b1}
Kuntimad G., Ranganath H.S. Perfect image segmentation using pulse coupled neural networks. IEEE Transactions on Neural Networks. 1999. V.10. N.3, P. 591--597.

\bibitem{b2}
Yide M., Qing L. Automated image segmentation using improved PCNN model based on cross-entropy. International Symposium on Intelligent Multimedia, Video and Speech Processing. 2004. P. 20--22.

\bibitem{b3}
Yide M., Chunliang Q. Study of Automated PCNN System Based on Genetic Algorithm. Journal of system simulation. 2006. V.18. N.3. P. 722--724.

\bibitem{b4}
Xiaodon G., Shide G., Daohen Y. A new approach for image segmentation based on unit-linking PCNN. Proceeding of the first International Conference on Machine Learning and Cybernetics. 2002. P.

\bibitem{b5}
Stewart R.D.F., Opper M. Kegion growing with pulse-coupled neural networks: An alternative to seeded region growing. IEEE Transactions on Neural Networks. 2002. V. 13. N. 6. P. 1557--1562.

\bibitem{b6}
Qing Z., Guanhui Y., Tingling G., Hong Z., Junxiao L. Fabric Defect Segmentation Based on Region
Growing PCNN Model. Computer application and software. 2011. V. 28. N. 11. P. 171--175.

\bibitem{b7}
Ma H.-R., Cheng X.-W.  Automatic Image Segmentation with PCNN Algorithm Based on Grayscale
Correlation. International Journal of Signal Processing, Image Processing and Pattern Recognition. 2014. V. 7. N. 5. P.249-258

\bibitem{b8}
Privalov M.V., Skobtsov Yu.A., Kudriashov A.G. Applying of neural network methods of
classification to segmentation of computer tomograms. Vestnik Khersonskogo natsional’nogo
technicheskogo universiteta [Herald of Kherson National Technical Univ.], 2010, no. 2,
pp. 103-109 (in Russian).

\bibitem{b9}
Chandhok C. A Novel Approach to Image Segmentation using Artificial Neural Networks and K-Means Clustering. International Journal of Engineering Research and Applications (IJERA). 2012. V. 2. N. 3. P. 274-279

\bibitem{b10}
Xu B., Lin S. Automatic Color Identification in Printed Fabric Images by a Fuzzy Neural Network. Computer Journal of AATICC Review. 2002. V.2. N.9. P. 42--45.

\bibitem{b11}
Yao K., Mignotte M., Collet C., Galerne P., Burel G. Unsupervised Segmentation Using a Self
Organizing Map and a Noise Model Estimation in Sonar Imagery. Computer Journal of Pattern Recognition Letters. 2000. V.33. N.9. P.1575--1584.

\bibitem{b12}
Aria E., Saradjian M., Amini J., Lucas C. Generalized Cooccurence Matrix to Classify IRS-1d
Images Using Neural Networks. Proceedings of ISPRS Congress. 2004. P. 117-123..

\bibitem{b13}
Awad M., Chehdi K., Nasri A. Multi Component Image Segmentation Using Genetic Algorithm and
Artificial Neural Network. Computer Journal of Geosciences and Remote Sensing Letters. 2007. V.4. N.4. P. 571--575.

\bibitem{b14}
Zhou Z., Wei S., Zhang X., Zhao X. Remote Sensing Image Segmentation Based on Self Organizing Map
at Multiple Scale. Proceedings of SPIE Geoinformatics: Remotely Sensed Data and Information. 2007.
P.122--126.

\bibitem{b15}
Kurnaz M., Dokur Z., Olmez T. Segmentation of Remote Sensing Images by Incremental Neural
Network. Computer Journal of Pattern Recognition Letters. 2005. V.26. N.8. P. 1096--1104.

\bibitem{b16}
Belim S.V., Kutlunin P.E. Boundary extraction in images using a clustering algorithm. Computer Optics. 2015. V. 39. N.1. P. 119-124. (In Russ.)

\bibitem{b17}
Belim S.V., Larionov S.B. An Algorithm of image segmentation based on community detection in graphs. Computer Optics. 2016. V. 40. N. 6. P. 904-910. (In Russ.)

\end{thebibliography}
\end{document}